%% file: main.tex
\pdfoutput=1

\documentclass[11pt]{article}

\usepackage[final]{acl}

\usepackage{times}
\usepackage{latexsym}

\usepackage[T1]{fontenc}

\usepackage[utf8]{inputenc}

\usepackage{microtype}

\usepackage{inconsolata}

\usepackage{graphicx}
\usepackage{hyperref}
\usepackage{color}
\usepackage{multirow}
\usepackage{booktabs}
\usepackage{makecell}
\usepackage{graphicx}
\usepackage{amsthm,amsmath,amssymb}
\usepackage{mathrsfs}
\usepackage{pifont}
\usepackage{natbib}
\usepackage{enumitem}
\usepackage{algorithm}
\usepackage{algorithmic}
\usepackage[table,xcdraw]{xcolor}
\usepackage{array}
\usepackage{tikz} 
\usepackage{tcolorbox}
\usepackage{listings}
\usetikzlibrary{mindmap,trees}
\usepackage{relsize}
\usepackage{subcaption}
\usepackage{amsfonts}
\usepackage{siunitx} 

\definecolor{LakeBlue}{RGB}{0,61,153}

\newcommand{\modelname}{SEAGraph}

\title{
\raisebox{-0.4cm}{
\includegraphics[width=1.2cm]{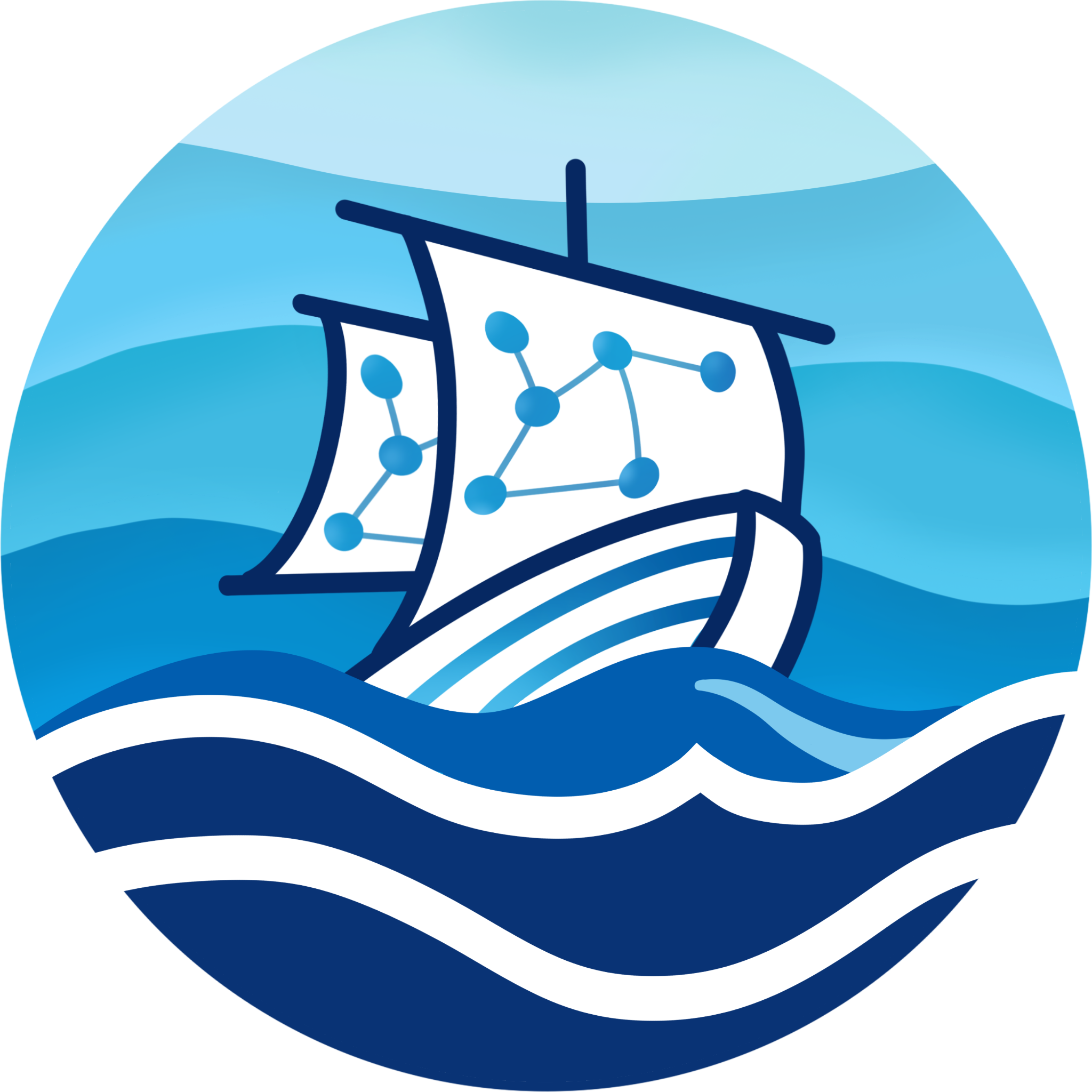}}
~
SEAGraph: Unveiling the Whole Story of Paper Review Comments
}

\author{Jianxiang Yu\thanks{~Equal Contribution},
Jiaqi Tan\footnotemark[1], 
Zichen Ding,
Jiapeng Zhu, 
{\bf Jiahao Li,} \\ 
{\bf Yao Cheng,} 
{\bf Qier Cui,} 
{\bf Yunshi Lan,} 
{\bf Yao Liu,}
{\bf Xiang Li\thanks{~~Corresponding author}} \\
East China Normal University,
Shanghai, China \\
\href{mailto:sea.ecnu@gmail.com}{sea.ecnu@gmail.com}
\\
\url{https://github.com/ecnu-sea/SEAGraph}
}

\begin{document}
\maketitle
\begin{abstract}

Peer review, as a cornerstone of scientific research, ensures the integrity and quality of scholarly work by providing authors with objective feedback for refinement.
However, in the traditional peer review process, 
authors often receive vague or insufficiently detailed feedback, which provides limited assistance and leads to a more time-consuming review cycle.
If authors can identify some specific weaknesses in their paper, they can not only address the reviewer's concerns but also improve their work.
This raises the critical question of how to enhance authors' comprehension of review comments.
In this paper, we present~\modelname, 
a novel framework developed to clarify review comments by uncovering the underlying intentions behind them.
We construct two types of graphs for each paper: the semantic mind graph, which captures the authors' thought process, and the hierarchical background graph, which delineates the research domains related to the paper.
A retrieval method is then designed to extract relevant content
from both graphs, facilitating coherent explanations 
for the review comments.
Extensive experiments show that~\modelname~excels in review comment understanding tasks, offering significant benefits to authors. 
By bridging the gap between reviewers' critiques and authors' comprehension, 
\modelname~contributes to a more efficient, transparent and collaborative scientific publishing ecosystem.
\end{abstract}

\input{section/introduction}
\input{section/relatedwork}
\input{section/method}
\input{section/experiment}

\section{Conclusion}
In this paper, we present~\modelname,
a novel framework designed to bridge the gap between reviewers’ comments and authors’ understanding. 
By constructing two distinct graphs—the semantic mind graph, which captures the authors' thought process, and the hierarchical background graph, which encapsulates the research background—the framework effectively models the context of a reviewed paper. 
The well-designed retrieval method ensures that relevant content from both graphs is used to generate 
coherent and 
logical 
explanations for review comments.
\modelname~not only enhances the clarity of reviews 
but also empowers authors to understand reviewer concerns more effectively, 
improving the quality of academic publications.

In a nutshell, we sincerely hope that our work does empower authors to not only gain a deeper understanding of reviews feedback but also elevate the quality of their papers, ultimately expediting both the advancement of research and the efficiency of the submission process.
We hope this work can inspire further efforts toward making peer review more accessible, informative, and impactful.

\clearpage

\section*{Limitations}

\modelname~is designed to assist authors in comprehending review comments during the peer review process, 
with a particular emphasis on the review-comment understanding stage. 
Here we elaborate on some of these constraints, along with intriguing future explorations.

\paragraph{Rebuttal mechanism.}
The rebuttal mechanism, where authors respond to reviewers' concerns and engage in further discussion, 
also plays a critical role in improving the paper~\cite{jin2024agentreview}. 
The success of multi-agent systems in executing complex tasks presents a promising opportunity~\cite{autogen, oscopilot}. 
In future research, we will explore simulating the rebuttal process through multi-agent communication, 
aiming to further bridge the understanding gap between reviewers and authors in papers and comments, 
thus advancing the rebuttal mechanism.

\paragraph{Enhancing Pipeline Stability.}
As an integrated pipeline,
\modelname~involves a relatively complex process. 
Certain components, 
such as the processes of searching for and downloading relevant papers, 
are dependent on network conditions.
To address this, 
we aim to continuously refine and optimize the underlying code, 
ensuring the robustness and stability of these technical operations while improving their overall efficiency.

\paragraph{More granular Evaluation.}
Although human evaluation and GPT-based evaluation can reflect the strengths and weaknesses of a model to some extent, they often involve significant subjectivity and lack consistency. This issue becomes particularly pronounced in open-ended generation tasks, where differences in standards and preferences among evaluators may lead to inconsistent results. 
Therefore, establishing a comprehensive, unified, and reproducible quantitative standard is crucial for more objective and fair assessment of model performance. Such a standard not only helps to minimize the influence of human bias but also provides more actionable feedback for subsequent model optimization and improvement.
In addition, a comprehensive evaluation requires more data and human involvement, which brings with it significant costs.

\paragraph{Challenges in Benchmarking.}
We plan to establish a standardized benchmark for the field of peer review in the future to help standardize and advance research practices in this area.
Currently, platforms such as OpenReview provide a wealth of publicly available papers and review data, offering valuable resources for studying the mechanisms and effectiveness of peer review. 
However, there are certain limitations to the use of these data.
On one hand, there is the issue of data incompleteness. Typically, an article will have three or four reviewers, or even more, and extracting and merging the key information from these reviews poses a significant challenge. 
Meanwhile, for review comments understanding task, 
the ground truth needs to explicitly capture the reasoning behind each reviewer’s comments and form a complete logical chain. 
This process not only requires the flexible use of LLMs for assistance but also demands the deep involvement of experts, which brings with it significant and potentially immeasurable costs.
On the other hand,
there are concerns regarding privacy protection.
Although the identities of authors are ostensibly anonymized, the peer review process allows senior roles such as Program Chairs and Area Chairs to access the actual identities of both authors and reviewers. 
This potential breach of privacy may pose challenges to the objectivity, fairness, and ethical considerations of related research.

\section*{Ethics Statement}
This work seeks to assist authors in better comprehending review comments.
We do not intend to suggest that some reviews are inherently of low quality or unhelpful. 
Instead, we appreciate that clearer and more comprehensible review comments can more effectively fulfill the primary objective of the peer review process—namely, 
to offer objective evaluations and constructive feedback aimed at improving the manuscript.
Recently, some academic conferences have introduced AI-assisted review bots to standardize reviewers' feedback. 
Through this work, we aim to benefit authors by enhancing their understanding of review comments, 
while also encouraging reviewers to consider the clarity of their feedback and strive for higher-quality reviews. 
Ultimately, we seek to foster a healthier and more harmonious academic interaction environment.

\section*{Acknowledgments}
This work is supported by National Natural Science Foundation
of China (No. 62202172).

\bibliography{custom}

\clearpage

\appendix

\input{section/appendix}

\end{document}

%% file: section/introduction.tex
\section{Introduction}
\begin{figure}
    \centering
    \includegraphics[width=0.5\textwidth]{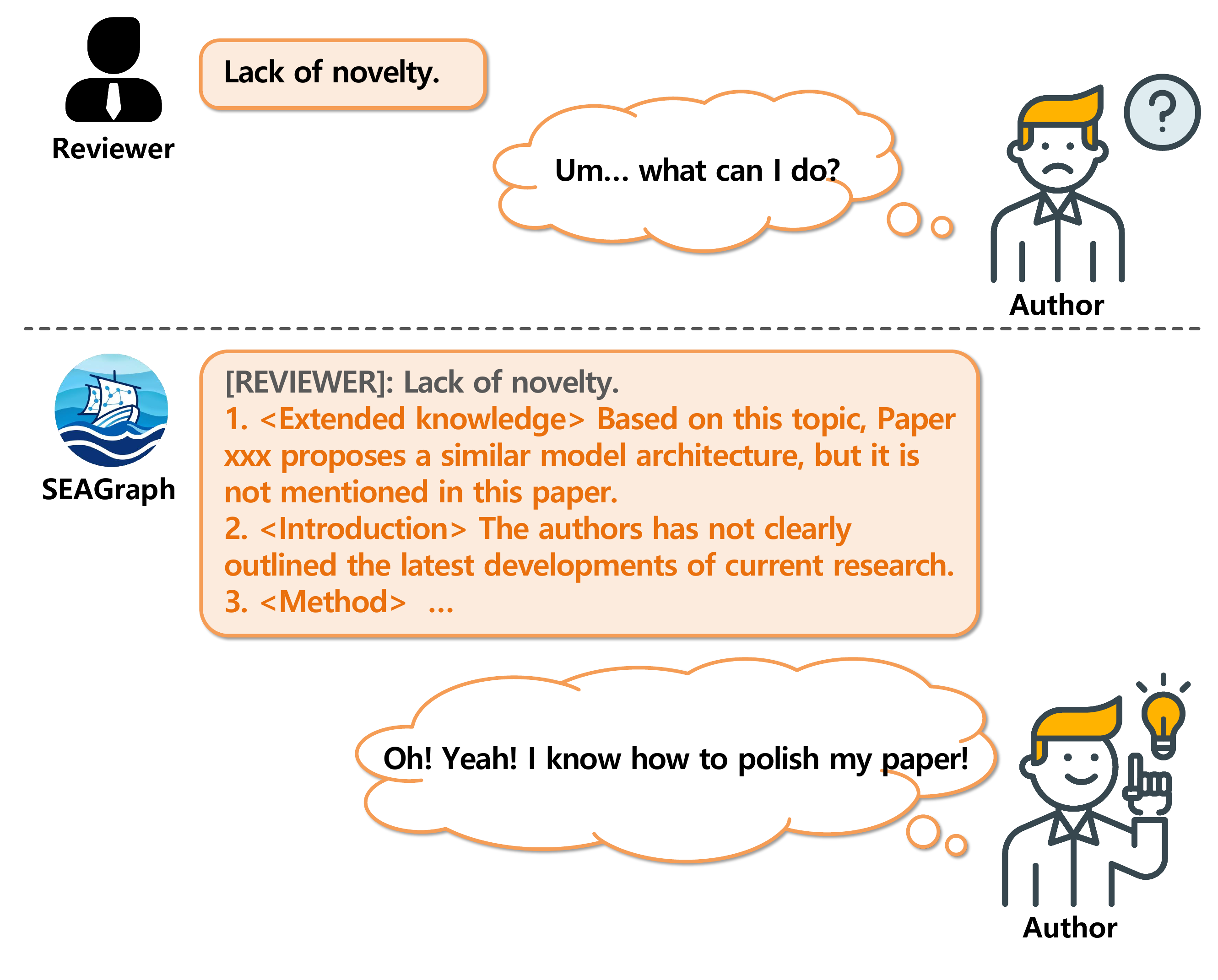}
    \caption{\modelname~ can help authors better understand reviewers’ comments by providing detailed insights and evidence. 
    }
    \label{fig:intro1}
    \vspace{-1em}
\end{figure}

In recent years, the number of academic publications has grown exponentially, creating a vast ``sea of papers''~\cite{bornmann2015growth,lin2023moprd}. 
Traditionally, authors rely on the peer review process to receive feedback on their manuscripts~\cite{lee2013bias,bjork2013publishing}.
The peer review process typically lasts for several months or even longer~\cite{horbach2018changing}, yet the crucial rebuttal phase - the limited window for author-reviewer communication - remains disproportionately short. During this brief interaction period, authors and reviewers must navigate complex technical discussions through written exchanges alone, without the benefit of face-to-face dialogue that could facilitate mutual understanding~\cite{verma2022lack}.
This time-constrained communication may leads to suboptimal outcomes: while reviewers may provide valuable insights, authors struggle to fully comprehend or effectively address these comments within the tight rebuttal timeframe. Therefore, when authors better understand reviewers' perspectives, it not only leads to improved revisions but also makes the entire review process more productive and rewarding for both authors and reviewers
(see Figure~\ref{fig:intro1}).

Currently, Large Language Models (LLMs) have shown powerful text comprehension and generation capabilities~\cite{achiam2023gpt, wei2022emergent}, offering new directions for
\emph{revealing
the underlying intentions behind each review comment}.
A straightforward approach is to provide LLMs with both the comment and the corresponding paper. 
Yet, it is usually difficult to feed an entire paper into
LLMs for identifying key points, 
as review comments typically focus on specific aspects rather than the entire paper.
Another alternative approach is using RAG (Retrieval-Augmented Generation)~\cite{cheng2024lift,jiang2023active}, which enhances reasoning by retrieving the most relevant passages from lengthy texts based on the query.
Nevertheless, the information retrieved by RAG tends to be fragmented, lacking clear 
logic~\cite{cao2024lego}.
In contrast, review comments are given based on the coherent logical structure formed when reviewers read the paper, which cannot be easily captured by fragmented segments.
Recently, 
the success of GraphRAG~\cite{graphrag},
which splits lengthy texts into discrete chunks and hierarchically connects them,
has inspired new directions.
Similarly, 
papers are inherently structurally organized with sections and subsections provided.
Therefore, we can format papers as structured graphs, from which logical chains can be extracted to facilitate a deeper understanding of review comments.

In this paper, we propose~\modelname, a novel framework designed to uncover the intentions behind paper reviews and enhance the understanding of review comments.
We construct two distinct graphs for each reviewed paper: \emph{a semantic mind graph} and
\emph{a hierarchical background graph}.
Building upon the principles of the mind map~\cite{DANTONI20062}, which employs visuospatial orientation to integrate information, we introduce the semantic mind graph to facilitate deeper semantic connections and organization of key knowledge points.
In addition,
the hierarchical background graph connects various related papers based on the themes of the paper, thereby simulating its research context. 
After the construction of the two graphs, 
we design a tailored retrieval method to extract the most relevant content from both graphs in response to each review comment. 
The extracted content is subsequently fed into LLMs to generate \emph{coherent and logical arguments} that explain the reviewer’s comments.
Overall, our contributions are summarized as follows:
\begin{itemize}[leftmargin=*]

\item We introduce a novel framework~\modelname, which pioneers the field of review comment understanding.

\item We construct a semantic mind graph and a hierarchical background graph for a 
paper, 
capturing its deep semantics and related domain knowledge.

\item We conduct extensive experiments to validate the effectiveness of our framework, which can help authors improve the quality of their papers.

\end{itemize}

\begin{figure*}
    \centering
    \includegraphics[width=0.98\linewidth]{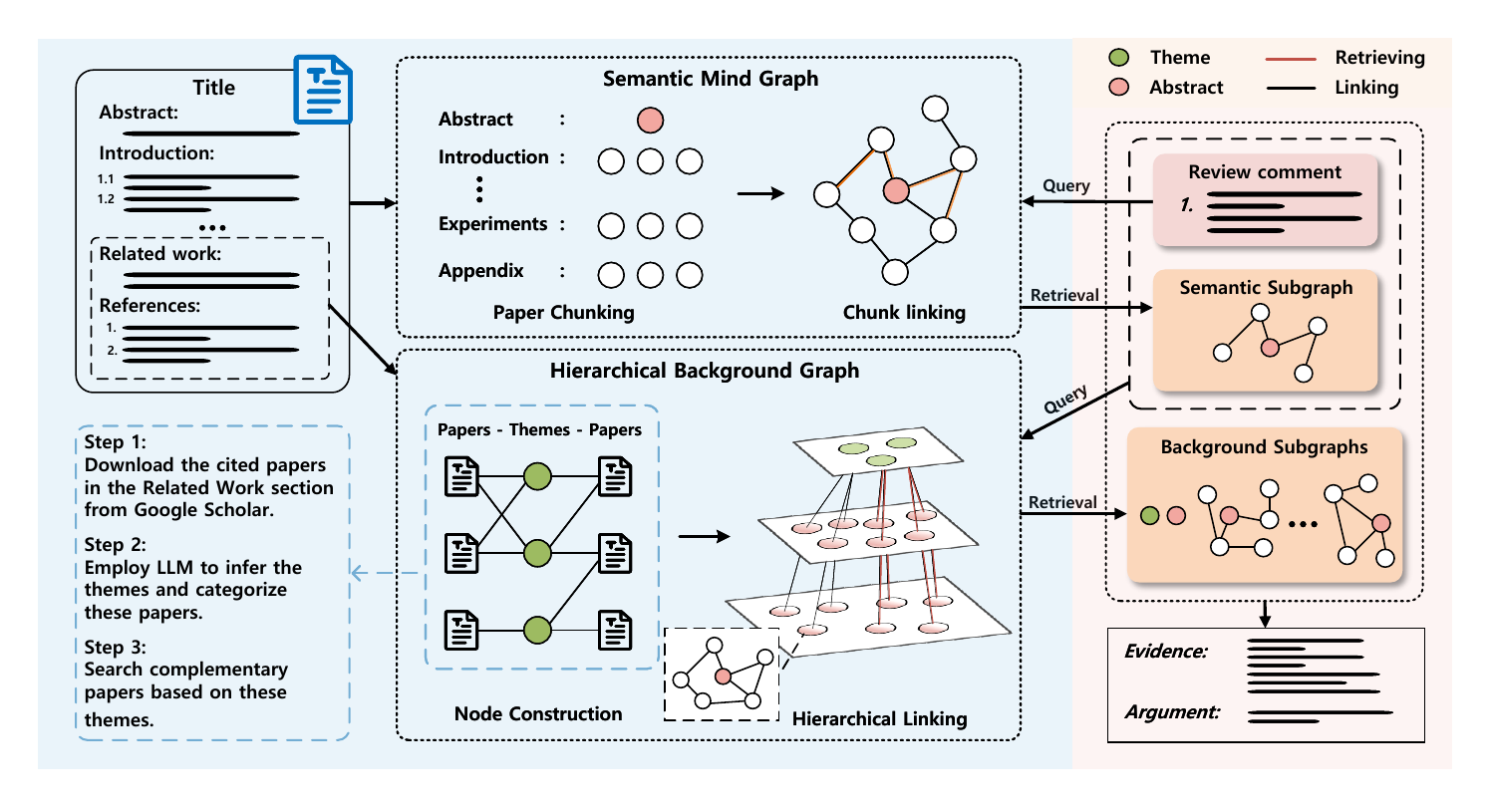}
    \caption{The overall framework of~\modelname~consists of the construction of the semantic mind graph and the hierarchical background graph, along with the corresponding retrieval module. The final retrieved content is fed into LLMs for review comment understanding. }
    \label{Framework}
\end{figure*}

%% file: section/relatedwork.tex
\section{Related Work}
\subsection{Retrieval-Augmented Generation}
RAG (Retrieval-Augmented Generation)
improves the generation performance of LLMs by incorporating external knowledge~\cite{lewis2020retrieval}.
Initially, naive RAG approaches follow a process including indexing, retrieval, and generation~\cite{li2022survey}.
The indexing phase involves cleaning and segmenting raw data into text chunks and encoding them into vector space. Retrieval and generation follow by encoding user queries, matching them with nearest chunks, and synthesizing context-aware responses using retrieved content.
Advanced RAG frameworks focus on enhancing the retrieval quality during the pre-retrieval and post-retrieval phase
like query rewriting, query expansion, and chunk reranking
~\cite{ma2023query,peng2024large,zheng2023take}.
Furthermore, some modular RAG approaches put forward new modules or pipelines to enhance the retrieval capability and alignment with task-specific requirements
\cite{yu2022generate,shao2023enhancing}.
Despite these advancements, RAG faces challenges in handling query-focused summarization tasks when queries target entire text corpora~\cite{cao2024lego}.
GraphRAG emerges as an innovative solution to address this challenge~\cite{peng2024graph}. 
\citet{graphrag} establish logical relationships between segments by connecting chunks or communities through a hierarchical structure.
\citet{wu2024medical} and \citet{sepasdar2024enhancing} 
construct specialized knowledge graphs, extending GraphRAG to the medical and soccer domains.
In this work,
we construct two logically connected graphs for the paper, leveraging the strengths of GraphRAG to better address the review comment understanding tasks.

\subsection{Large Language Models in Peer Review}
Recently,
LLMs have made remarkable progress in text generation tasks~\cite{zhao2023survey,ouyang2022training,lbs3}, 
prompting researchers to explore new opportunities in the field of peer review ~\cite{li2024sentiment,checco2021ai}.
A significant focus has been placed on generating automated reviews to enhance the quality of academic papers~\cite{reviewer2}.
For example, \citet{liu2023reviewergpt} and~\citet{liang2023can} customize prompts to guide GPT-4 in generating scientific feedbacks,
while \citet{yu2024automated} and \citet{academicgpt} refine LLMs through fine-tuning and pretraining.
Expanding on this,
\citet{jin2024agentreview} simulate the review process with LLMs to analyze evaluation factors.
Meanwhile,
\citet{ye2024we} highlight risks in automated peer review, and \citet{yu2024your} explore challenges in distinguishing human and LLM-generated reviews.
Besides,
previous studies have highlighted various challenges in the peer review and rebuttal process.
\citet{kuznetsov2024can} note the time-intensive nature of peer review, requiring extensive discussions, while \citet{purkayastha2023exploring} emphasize challenges in rebuttals due to language barriers and experience gaps. \citet{huang2023makes} stress the need for rebuttals to address all reviewer concerns and reach consensus. These findings highlight the importance of improving authors’ understanding of feedback.
Our work aims to leverage LLMs to understand the intent of review comments, thereby assisting authors in polishing their papers.

%% file: section/method.tex
\section{\modelname}
Accurately simulating the perspective of a reviewer necessitates not only enabling LLMs to understand the content of the paper, but also equipping them with the background knowledge required for peer review. 
To this end, we design a \emph{Semantic Mind Graph} and a \emph{Hierarchical Background Graph} to model the two corresponding types of knowledge structures.
The former captures the paper's intrinsic arguments and evidence, while the latter formalizes the broader domain context essential for an informed critique. 
This explicit, dual-graph approach significantly aids LLMs in integrating these different knowledge types, thereby fostering a more robust and interpretable emulation of a reviewer's reasoning.
In the following, we present the details of 
each module in \modelname, with the overall framework illustrated in Figure~\ref{Framework}.

\subsection{Data Preprocessing}
\paragraph{Paper Processing.}
Our dataset consists of the PDF versions of academic papers and their corresponding review comments. 
We begin by utilizing Nougat~\cite{nougat} as the parser, a model built on the Visual Transformer architecture specifically tailored for extracting information from academic documents.
Then, we construct the \textbf{S}emantic mind graph and the \textbf{H}ierarchical background graph for each reviewed paper, denoted as \( \mathcal{G}_S(\mathcal{V}_S, \mathcal{E}_S) \) and \( \mathcal{G}_H(\mathcal{V}_H, \mathcal{E}_H) \), where \( \mathcal{V} \) and \( \mathcal{E} \) represent nodes and edges, respectively. 

\paragraph{Review Comments Extraction.}
We input the entire review into LLM to extract individual comments 
like 
``\texttt{Strengths}, ``\texttt{Weaknesses}'', and ``\texttt{Questions}.''
Then,
all 
review comments 
are defined as a query set \( \mathcal{Q} \), where \( q \in \mathcal{Q} \) represents a single comment.

\subsection{Semantic Mind Graph Construction}
\label{sec:smg_construction}
A paper is structurally organized into different sections, 
while key points of the paper may be scattered across various parts. 
The entire paper can be structured like a mind graph, where content progressively branches out from different paragraphs.
Our goal is to construct a semantic mind graph to model the writing logic of a paper.
We next detail the main steps.
\paragraph{Paper Chunking.}
We first use the Spacy library~\cite{spacy2} to break a full paper down to sentence level, allowing us to assess the relevance between sentences and decide whether adjacent sentences should be merged into chunks.
Specifically,
we utilize Sentence-BERT~\cite{sentence-bert} to encode both sentences and chunks,
and 
design a semantic relevance measure to 
determine whether the current chunk is related to the next sentence. 
Subsequently,
we place the first sentence $s_1$ into the initial chunk $c_{\text{current}}$.
For each subsequent sentence $s_i$, we compute the embedding similarity between $s_i$ and $c_{\text{current}}$.
If the similarity exceeds a threshold \( \theta_1 \), we merge the sentence into the current chunk by:
\begin{equation}
\nonumber
c_{\text{current}} \leftarrow c_{\text{current}} \cup s_i \quad \text{if } \mathit{f}(h_{s_i}, h_{c_{\text{current}}}) \geq \theta_1,
\end{equation}
where \( h_{s_i} \) and \( h_{c_{\text{current}}} \) represent the embeddings of the \( i \)-th sentence and the current chunk, respectively.
Here, \(\mathit{f}(\cdot,\cdot)\) is the similarity function (e.g. cosine similarity).
If the similarity is less than the threshold,
we end the current chunk and start a new one: \( c_{\text{new}} \leftarrow s_i \).
Then we repeat the above steps to divide a paper into chunks.
Additionally,
a maximum chunk size is also set to prevent excessive imbalance in the length of different chunks.
Finally, 
the chunk nodes of the paper can be represented as \( \mathcal{V}_S = \{c_a, c_1, c_2, \dots c_n\}\), where \( c_a \) denotes the abstract node of the paper.
In this way,
the content within the same chunk is closely related. 

\paragraph{Chunk Linking.}
Given the segmented chunks, the next step is to link them based on their
contextual and
semantic relationships.
Generally, chunks within the same (sub)section often share the same topic. 
For example, the ``Method'' section describes the paper's research methodology, 
while the ``Experiments'' section validates the proposed method through experimental design. 
Therefore,
adjacent chunks in a sequential order are probably highly correlated and we call this \emph{contextual correlation}.
We can establish connections between them by setting
$e_{c_a,c_1} = 1$ and
$e_{c_i,c_{i+1}} = 1, \forall i \in [1, n-1]$
where \(e\) denotes the edge between two chunk nodes.
This approach also helps mitigate issues that sentences with high semantic relevance may be split into different chunks due to chunk size constraints.

\begin{figure}[t]
    \centering
    \includegraphics[width=1.0\linewidth]{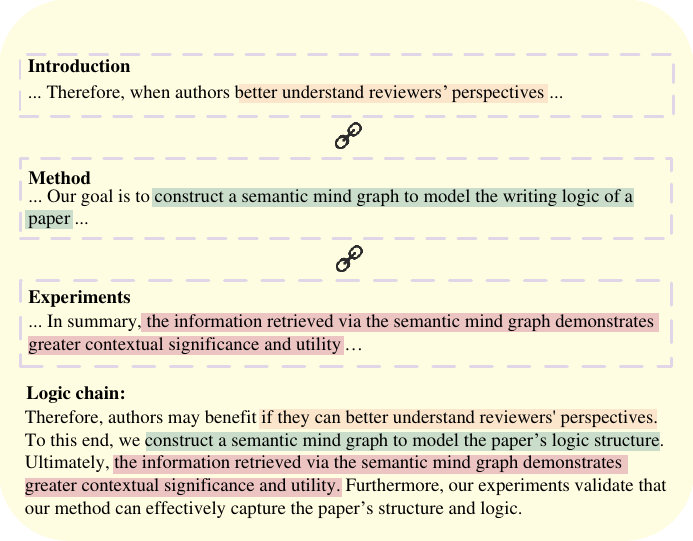}
    \caption{Chunks from different sections can form a coherent logic chain. 
    }
    \label{fig:smg}
\end{figure}

Further, the authors may not simply follow a linear mind in organizing the paper.
As shown in Figure~\ref{fig:smg}, 
the ``Introduction'' section of a paper often lays the groundwork for understanding the problem, which is further elaborated in the ``Method'' section with detailed descriptions of the proposed approach or framework. 
Subsequently, the ``Experiments'' section validates these methods through practical evaluations. 
If these segments are extracted individually, they can still form a coherent logic. We call this \emph{semantic correlation}. 
To capture their correlations, 
we compute the semantic similarities between different chunks and set a threshold $\theta_2$ to connect highly similar chunks:
\begin{equation}
    e_{c_i,c_j} = 1 \quad \text{if } \mathit{f}(h_{c_i}, h_{c_j}) \geq \theta_2
\end{equation}

Finally, the paper is transformed into a
semantic mind graph,
where linked chunks represent 
either contextual proximity or semantic similarity.

\subsection{Hierarchical Background Graph Construction}
To effectively review a paper, a reviewer not only needs a deep understanding of the content but also a solid grasp of the knowledge in 
corresponding fields.
Therefore,
we construct a background graph with hierarchical relationships to simulate reviewers' domain knowledge.
This graph is organized into a three-layer structure: 
the themes of the reviewed paper, the abstracts of relevant papers, and the semantic mind graph for each individual paper.

\paragraph{Cited Paper Search.}
We first locate the cited papers in the ``Related Work'' section of the reviewed paper,
which
represent the existing research achievements in the field.
Subsequently,
we extract the publication details of these papers from the ``Reference'' section. 

\paragraph{Theme Summarization.}
We next crawl the PDFs of referenced papers from Google Scholar, 
parse them into markdown format, and extract the abstracts and titles of each paper.
Then,
we feed them into LLM to summarize multiple themes and assign corresponding papers to each theme.
In this way, we obtain a theme set related to the reviewed paper, denoted as $ \mathcal{T} = \{t_1, t_2, \dots, t_m\}$ and $t$ refers to the descriptive summarization of a theme node.

\paragraph{Complementary Papers.}
The authors may not always reference all foundational or cutting-edge papers in the field. Therefore, we aim to enrich the paper by incorporating both fundamental and recent studies within the research domain.
Based on the extracted themes, we search and crawl the most popular and recent papers related to these themes from Google Scholar
to enrich the background graph in terms of breadth and timeliness.
After identifying the relevant papers related to the paper, we apply the method from Section~\ref{sec:smg_construction} to construct a semantic mind graph for each of them. 

\paragraph{Hierarchical Linking.}
For a reviewed paper,
we construct its hierarchical background knowledge graph based on theme nodes, abstract nodes, and semantic mind graphs.
The first level includes multiple theme nodes, each corresponding to a thematic description that encapsulates the research topics.
The second level connects these theme nodes to abstract nodes, where each abstract serves as a concise summary of a paper, 
representing its key ideas and maintaining a direct association with its respective theme. 
The third level extends from the abstract nodes to semantic mind graphs, which provide fine-grained information, offering a deeper insight into the paper’s structure and details. 
This hierarchical design clearly delineates the logical progression from themes to papers and further to detailed information, forming a systematic framework for representing the research background.

\subsection{Semantic Mind Graph Retrieval}
\label{sec:smg_retrieval}
We next introduce retrieving the semantic mind graph based on review comments and obtaining the relevant supporting texts.
Given a review comment as a query, we first calculate the probability distribution of the query over the semantic mind graph by calculating the textual similarity between the query and each chunk node $c_j$:
\begin{equation}
    P(c_j) = \frac{\mathit{f}(h_q, h_{c_j})}{\sum_{i=1}^{n} \mathit{f}(h_q, h_{c_i})},
\label{eq:prob}
\end{equation}
where \( P(c_j) \) represents the probability distribution over the nodes in the semantic mind graph,
\(\mathit{f(\cdot,\cdot)}\) is the similarity function and \(h_q\) denotes the embedding of the query.

Then we iteratively retrieve chunk nodes that can help explain the review comments.
We start by randomly sampling
$k$ chunk nodes based on the probability computed in Eq.~\ref{eq:prob} and add them to an empty node set $\bar{\mathcal{V}}$. 
After that,
we explore the one-hop neighbors of these newly sampled chunk nodes and add them to $\bar{\mathcal{V}}$. 
Given a chunk node $c_i \in \bar{\mathcal{V}}$,
suppose chunk node $c_j$ as its one-hop neighbor and we calculate:
\begin{equation}
    \text{score}_i(h_{c_j}) = \alpha \cdot P(c_j) +  \mathit{f}(h_{c_i}, h_{c_j}),
    \label{eq:nei_prob}
\end{equation}
where \(\mathit{f}(h_{c_i}, h_{c_j})\) denotes the cosine similarity between 
the embeddings of the two chunks and $\alpha$ is a hyperparameter to control term balance.
In Eq.~\ref{eq:nei_prob},
the score is used to measure the relevance between query and chunk $c_j$.
In particular,
the second term calculates the similarity between chunk nodes $c_i$ and $c_j$,
which indirectly reflects the relation between query and chunk $c_j$ via chunk $c_i$.
After the scores are computed,
we select chunk nodes with the highest 
scores and further add them into $\bar{\mathcal{V}}$.
We repeat the above process to retrieve more chunk nodes.
Finally, all the chunk nodes in $\bar{\mathcal{V}}$ constitutes a subgraph that is relevant to the given review comment.

\subsection{Hierarchical Background Graph Retrieval}

To further mine the background knowledge related to a paper, 
we conduct an in-depth hierarchical background graph retrieval
based on the review comment and the 
corresponding semantic mind subgraph obtained in the previous section.
The hierarchical retrieval process refines from (1) \emph{theme level} to (2) \emph{abstract level}, and finally to (3) \emph{chunk level}, ensuring background knowledge obtained at different levels of granularity.
We apply 
scoring method in Eq.~\ref{eq:nei_prob} to the three levels of nodes and select the nodes with higher scores.

We first begin the retrieval process at the \emph{theme level}, aiming to extract themes related to a review comment. 
For example, 
if a reviewer questions \emph{whether the proposed method shares similarities with certain techniques in the fields of computer vision (e.g., contrastive learning in CV)},
we retrieve descriptions of related theme nodes to broadly align with the reviewer’s concerns.
Based on the theme-level information, we then proceed to the \emph{abstract-level} retrieval, focusing on papers related to the identified themes. 
Note that abstracts summarize key research questions, methodologies, and conclusions, providing a concise yet comprehensive overview. 
Therefore,
abstract-level retrieval is particularly useful for understanding review comments that compare the originality of the proposed approach with existing methods.

Finally, we step into the \emph{chunk-level} 
retrieval. 
Chunk nodes include detailed information, such as experimental setups and results.
They can be used to better understand review comments, thereby providing details to revise the paper.

After retrieval across all three granularity levels, the top-$k$ relevant nodes are ranked and selected to ensure that the most pertinent evidence is used for understanding the review comments.
More implementation details are provided in the Appendix~\ref{app:hkeg_equation}.

Upon retrieving informative nodes from both graphs, we feed the corresponding text along with the query into LLM to generate an explanation for the review comment.

%% file: section/experiment.tex
\section{Experiments}

\subsection{Experimental Setup}

\paragraph{Datasets.}
We collect a total of 1,256 review comments from ICLR submissions over the past five years via the OpenReview platform\footnote{\url{https://openreview.net/}}. 
Each comment contains no more than 200 characters, as longer reviews are typically considered sufficiently comprehensive and thus less suitable for our analysis.
The associated papers span diverse areas of artificial intelligence, with citation counts ranging from highly cited works to those with minimal impact, covering both recent studies and earlier publications.

\begin{figure}[t]
    \centering
    \includegraphics[width=0.75\linewidth]{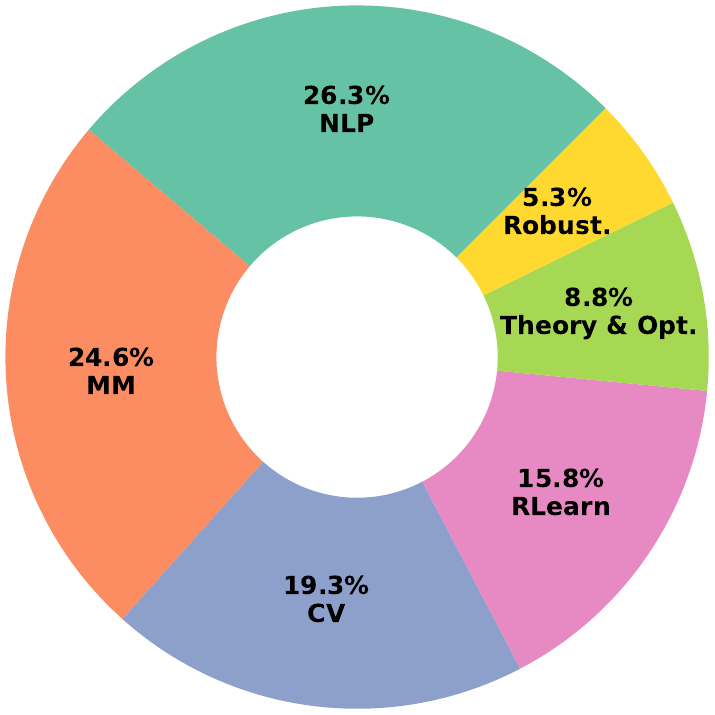}
    \caption{Research paper topic distribution across six key areas. NLP: Natural Language Processing; MM: Multimodal Learning; CV: Computer Vision; RLearn: Representation Learning; Theory\&Opt.: Theory and Optimization; Robust.: Robustness.
    }
    \label{fig:paper_cater}
\end{figure}

\begin{figure}[t]
    \centering
    \includegraphics[width=1.0\linewidth]{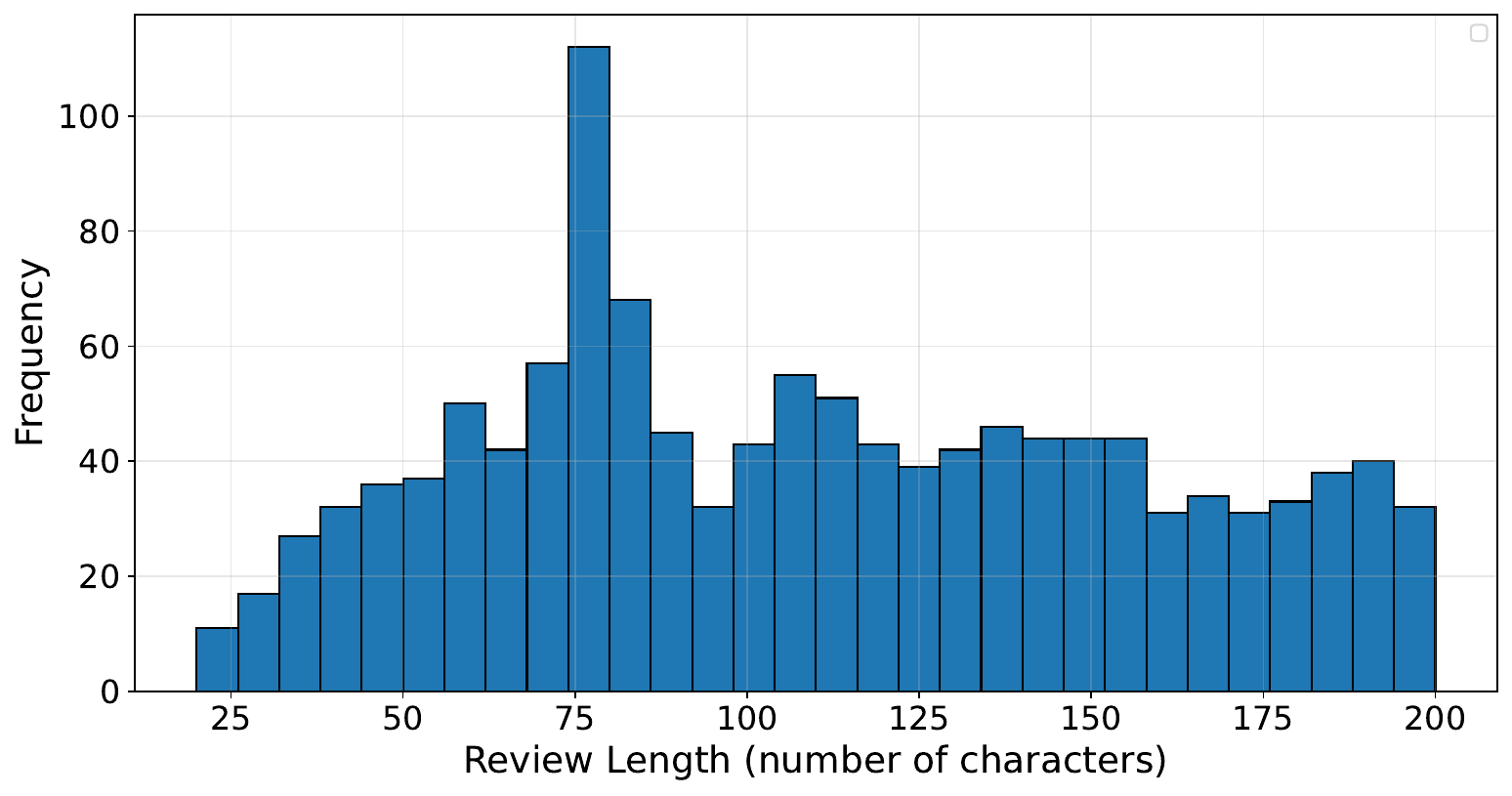}
    \caption{Distribution of Review Lengths.
    }
    \label{fig:review_length}
\end{figure}

The research papers are categorized into six major areas: Natural Language Processing (NLP), Multimodal Learning (MM), Computer Vision (CV), Representation Learning (RLearn), Theory and Optimization (Theory\&Opt.), and Robustness (Robust.). 
Figure~\ref{fig:paper_cater} illustrates the specific proportion of each category.
Further,
Figure~\ref{fig:review_length} presents the distribution of review lengths in our dataset. A prominent peak appears around 75 characters, indicating that a large number of short comments tend to concentrate at this length.

\begin{figure*}[t]
    \centering
    \includegraphics[width=1.0\linewidth]{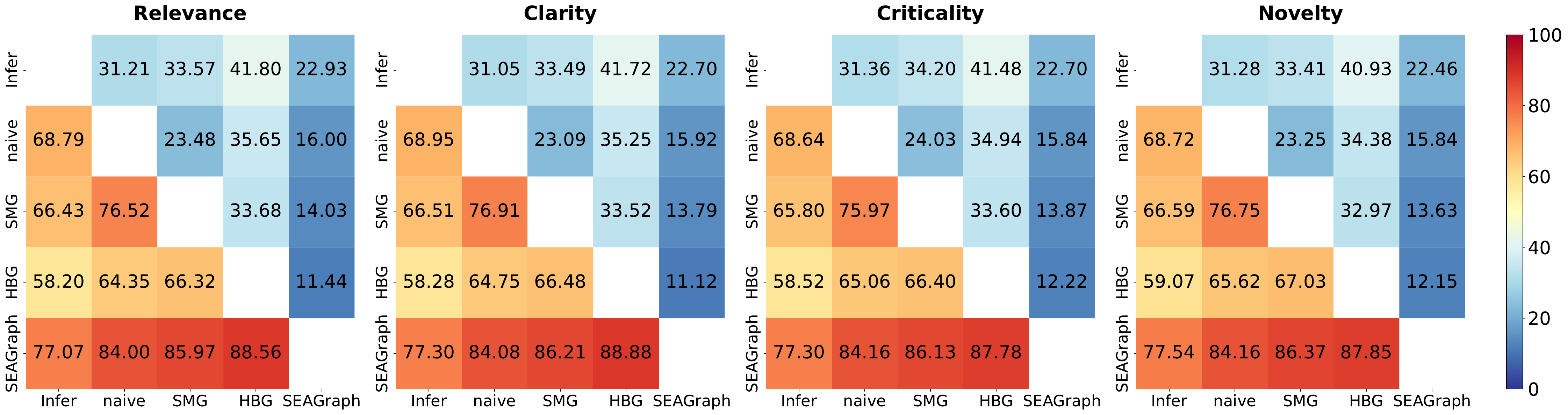}
    \caption{Automated evaluation results for Review Comments Understanding, measured by Win-Rate (\% $\uparrow$). }
    \label{fig:qwen_auto}
\end{figure*}

\paragraph{Baseline Methods.}
To validate the effectiveness of SEAGraph in terms of graph construction and retrieval, we compare it with the following two categories of baseline methods:
(1) Direct inference method: {\textbf{DirectInfer}} takes the review comment and the full parsed paper as input to directly reason about the understanding of each review comment.
(2) RAG-based methods: \textbf{RAG-naive} computes the similarity between each review comment and every chunk of the paper, selecting the top-$k$ chunks to combine with the review comment as input; 
\textbf{RAG-SMG} utilizes only the construction and retrieval of the \textbf{S}emantic \textbf{M}ind \textbf{G}raph; 
\textbf{RAG-HBG} relies solely the construction and retrieval of the \textbf{H}ierarchical \textbf{B}ackground \textbf{G}raph. 

We conduct experiments on all baseline models using the same open-source foundation models, \emph{Qwen3-8B}\footnote{\url{https://qwenlm.github.io/}} and \emph{Ministral-8B-Instruct-2410}\footnote{\url{https://mistral.ai/}}. 
Due to space constraints, the experimental details for Ministral are provided in Appendix~\ref{app:mistral}.
The example of SMG and HBG construction is in Appendix~\ref{app:construct_example} and the example of SEAGraph revealing review comments is in Appendix~\ref{app:generated_example_seagraph}.

\input{table/retrievel_result}
\input{table/generate_result_qwen}

\paragraph{Evaluation Protocol.}
Since the task of review comments understanding lacks a definitive ground truth and exhibits significant diversity in generated content, 
we design two evaluation methods: \emph{automated evaluation} and \emph{human evaluation}.
For automated evaluation, 
given the powerful text comprehension capabilities of LLMs to play as a judge~\cite{li2024generation}, we employ 
\emph{gpt-4o-2024-11-20}
and
\emph{Qwen3-14B}
as the evaluation model to provide objective judgments for the task.
For human evaluation, we recruit 40 experts from diverse academic background,
detailed information is provided in Appendix~\ref{app:experts}.

\subsection{Automated Evaluation Results.}
\label{sec:4.2}
\paragraph{Evaluation Metric.}
There are four main assessment metrics for human and automated evaluation: 
(1) \emph{Relevance}: Assesses the alignment between the provided evidence and the review comments.
(2) \emph{Clarity}: Evaluates how clearly and effectively the information is presented for ease of understanding.
(3) \emph{Criticality}: Examines the depth of analysis and the extent to which the feedback reflects constructive thought.
(4) \emph{Novelty}: Measures the inclusion of fresh insights or new evidence.
The specific meaning is provided in Appendix~\ref{specific_meaning_metrics}.

\paragraph{Result.}
In the automated evaluation, due to the excessive total text length generated by the five methods, we employ a pairwise ranking approach for assessment.
Figure~\ref{fig:qwen_auto} presents the results for review comments.
The values in the heatmap represent the win rate of the method shown on the vertical axis over the method on the horizontal axis. 
From the figure, 
we can see that:
(1) SEAGraph consistently outperforms all baseline methods across all four evaluation metrics
—Relevance, Clarity, Criticality, and Novelty—
demonstrating its superior ability to understand 
review comments. 
Its win rates exceed 77\% in all pairwise comparisons, with several cases reaching above 85\%, indicating a strong advantage in both content alignment and critical interpretation.
(2) RAG-SMG and RAG-HBG serve as moderately effective strategies, incorporating either semantic or hierarchical knowledge structures. 
Both methods contribute meaningful improvements, demonstrating that integrating information from distinct knowledge dimensions—internal semantics and external context—provides valuable support for interpreting review comments.
They show noticeable improvements over the naive and Infer baselines, suggesting that structured input plays an important role in enhancing the model’s comprehension of review intent.
(3) DirectInfer and RAG-naive methods perform poorly across all metrics,
particularly in tasks requiring deeper understanding such as clarity and novelty.
Their limitations stem from relying either solely on the original paper content or on unstructured retrieved evidence, both of which lack the logical organization needed to support reviewer-level reasoning.
Overall, 
\modelname~effectively generates insights for interpreting review comments.

\begin{figure*}[ht]
    \centering
    \includegraphics[width=0.96\linewidth]{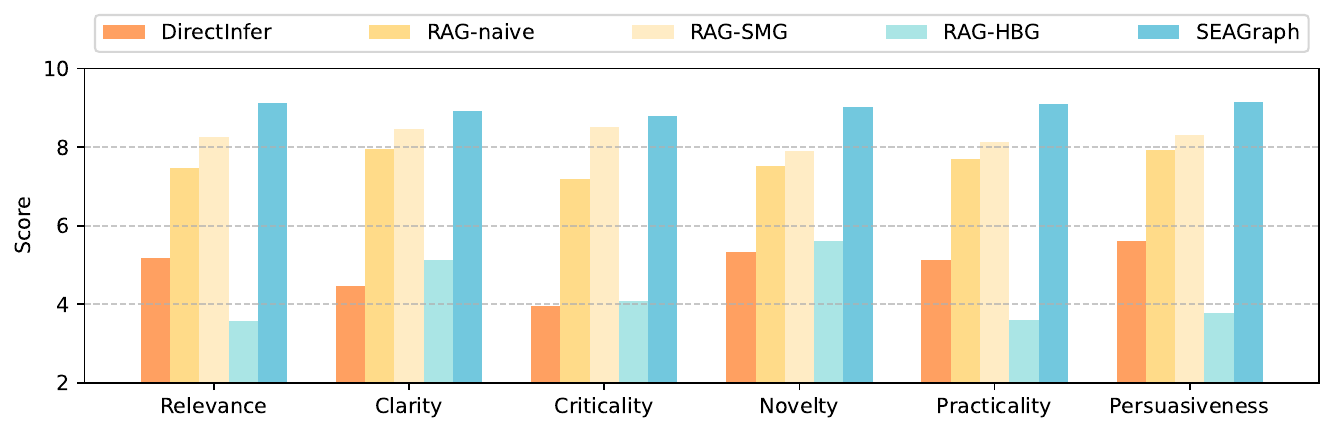
    }
    \caption{Human evaluation results for Key concerns in Reviews ($\uparrow$). }
    \label{fig:human_eval_exp3}
\end{figure*}


\subsection{Quantitative Evaluation of Retrieval}

The evaluation of retrieved content includes five distinct metrics:
(1) Relevance,
(2) Specificity,
(3) Novelty,
(4) Logic,
(5) Explainability.
The
detailed meanings are shown in Appendix~\ref{specific_meaning_metrics}.
Due to the extensive length of the retrieved content, 
both the content and corresponding review comment are directly input into gpt-4o for scoring. 
The results, as presented in the Table~\ref{tab:retrieval_evaluation}, demonstrate that RAG-SMG outperforms other methods in the majority of cases. 
Specifically, while RAG-SMG scores slightly lower than RAG-naive on the Relevance metric, this discrepancy may be attributed to the retrieval of chunk nodes with marginally lower semantic similarity through the logical structure of the semantic mind graph. 
Conversely, the external background knowledge retrieved by RAG-HBG exhibits limited relevance to the review comment, leading to consistently lower scores across various evaluation metrics. 
In summary, the information retrieved via the semantic mind graph demonstrates greater contextual significance and utility.
Additionally, we include retrieval examples in Appendix~\ref{app:retreiva_example} to further illustrate the advantages of the Semantic Mind Graph.

\subsection{Human Evaluation Results}
\label{sec:4.4}
In the human evaluation process, experts are invited to rank the review understanding results produced by five different methods based the following metrics.
We adopt the same four automatic evaluation metrics introduced in Section~\ref{sec:4.2}: Relevance, Clarity, Criticality, and Novelty.
Additionally, two customized human-centric metrics are tailored for evaluation:
(1) \emph{Persuasiveness}:  Focuses on the logical reasoning and the ability of arguments to persuade.
(2)	\emph{Practicality}: Gauges the usefulness and applicability of the information for authors.
The results are summarized in Table~\ref{tab:human_eval_generate_qwen}, where the scores indicate the average ranking over all samples.

We can observe that:
(1) \modelname~achieves the highest performance across all metrics, underscoring the effectiveness of our framework.
(2) In terms of average rankings, 
RAG-SMG performs better than RAG-naive, which in turn outperforms DirectInfer.
These findings strongly support our motivation that logically retrieved content significantly improves the ability of LLMs to understand review comments.
(3) Although RAG-HBG performs poorly—mainly because its retrieved content consists only of background knowledge without incorporating the internal knowledge of the paper—\modelname\ enhances the LLM’s ability to understand the paper by integrating external background knowledge on top of RAG-SMG, enabling it to extract more meaningful information for reasoning.
Consequently,
constructing the semantic mind graph and the hierarchical background graph can provide valuable support for understanding paper reviews from different perspectives.

\section{Analysis of Human Evaluation Consistency}
To assess the reliability and effectiveness of human evaluation
in our study, we analyze the NDCG@5 scores across Qwen3-8B.
NDCG@5 is computed based on independent human judgments for the model’s generated outputs.
In the context of human evaluation, this metric quantifies how well annotators can distinguish and rank high-quality model outputs from lower-quality ones within the model’s generated responses. 
By computing NDCG@5 scores across multiple evaluation criteria, we aim to evaluate both the overall reliability and the sensitivity of human judgments in capturing meaningful differences in response quality.

The results are summarized in Table~\ref{ndcg_qwen} and reflect average NDCG@5 scores across six evaluation criteria: Relevance, Clarity, Critical Insight, Novelty, Persuasiveness, and Practicality.
As shown in the table, 
Qwen3-8B achieves relatively high NDCG@5 scores across all evaluation criteria, indicating that human annotators were able to consistently rank the quality of its generated responses. These results suggest strong reliability and agreement among annotators, likely because the model produces stable and coherent outputs that reduce ambiguity during evaluation.

\input{table/ndgc_qwen}

\subsection{Key Concerns in Reviews}
In this section, we consider the issues highlighted in the review comments as the key concerns of the paper.
We first use LLMs to summarize all the reviews of each paper, identifying the concerns most frequently mentioned by the reviewers.
Then,
we employ six metrics consistent with those in Section~\ref{sec:4.4} 
and perform a human evaluation to quantify and compare the effectiveness of ~\modelname~in understanding and explaining these concerns with other methods.
The results,
shown in Figure~\ref{fig:human_eval_exp3},
indicate that \modelname~consistently achieves best scores across all evaluation metrics, 
particularly excelling in persuasiveness and practicality,
demonstrating strong alignment with human preferences.
RAG-SMG also shows strong performance in all metrics, as the concerns of reviewers are closely aligned with the content of the paper.
In contrast, 
RAG-HBG has a poor performance since the supplemental background knowledge is less helpful to the key concerns.
Overall,
both \modelname~and RAG-SMG show superior performance,
garnering higher recognition for their practicality and persuasiveness from human evaluators, 
further confirming the superiority of our framework in assisting authors to understand reviews.

%% file: table/retrievel_result.tex
\begin{table}[t]
    \centering
    \resizebox{1.0\linewidth}{!}{
    \begin{tabular}{lccc}
    \toprule
    \textbf{Metric $\uparrow$} & \textbf{RAG-naive} & \textbf{RAG-SMG} & \textbf{RAG-HBG} \\
    \midrule
    \textbf{Relevance}      & \textbf{7.59} & 7.53 & 5.33 \\
    \textbf{Specificity}    & 7.54 & \textbf{7.60} & 5.47 \\
    \textbf{Novelty}        & 6.46 & \textbf{6.54} & 4.76 \\
    \textbf{Logic}          & 7.65 & \textbf{7.68} & 5.23 \\
    \textbf{Explainability} & 7.06 & \textbf{7.09} & 4.61 \\
    \bottomrule
    \end{tabular}}
    \caption{Automated evaluation results for retrieval.}
    \label{tab:retrieval_evaluation}
    \vspace{-0.6em}
\end{table}

%% file: table/generate_result_qwen.tex
\begin{table*}[t]
    \setlength\tabcolsep{3pt} 
    \renewcommand{\arraystretch}{1.2} 
    \centering
    \resizebox{0.9\linewidth}{!}{
    \begin{tabular}{l|cccccc|c}
    \toprule
    \textbf{Method}  & \textbf{Relevance} & \textbf{Clarity} & \textbf{Criticality} & \textbf{Novelty} & \textbf{Practicality} & \textbf{Persuasiveness} & \textbf{Avg.Rank} \\
    \midrule
DirectInfer & 3.45 & 3.85 & 3.87 & 3.60 & 3.41 & 3.62 & 3.63 \\
RAG-naive  & 3.11 & 2.93 & 2.90 & 3.24 & 3.16 & 2.97 & 3.05 \\
RAG-SMG & \underline{2.61} & \underline{2.55} & \underline{2.37} & \underline{2.81} & \underline{2.69} & \underline{2.63} & \underline{2.61} \\
RAG-HBG  & 3.94 & 3.70 & 3.84 & 3.40 & 3.91 & 3.79 & 3.76 \\
SEAGraph  & \textbf{1.89} & \textbf{1.97} & \textbf{2.02} & \textbf{1.95} & \textbf{1.83} & \textbf{1.99} & \textbf{1.94} \\
    \bottomrule
    \end{tabular}
    }
    \caption{Human evaluation results for Review Comments Understanding (Rank-Based \textbf{$\downarrow$}). 
    We highlight the best score on each metric in \textbf{bold} and the runner-up score with an \underline{underline}.
    }
     \label{tab:human_eval_generate_qwen}
\end{table*}

%% file: table/ndgc_qwen.tex
\begin{table}[htbp]
  \centering
    \resizebox{0.62\linewidth}{!}{
  \begin{tabular}{lcc}
    \toprule
    Metric         & Qwen3-8B  \\
    \midrule
    Relevance         & 0.9105           \\
    Clarity           & 0.9066       \\
    Critical Insight  & 0.8991       \\
    Novelty           & 0.8991       \\
    Persuasiveness    & 0.8874        \\
    Practicality      & 0.8995     \\
    \bottomrule
  \end{tabular}
  }
  \caption{NDCG@5 of Qwen3 based on human judgments.}
  \label{ndcg_qwen}
\end{table}

%% file: section/appendix.tex
\section{Performance with Ministral-8B}
\label{app:mistral}
In this section, we evaluate a subset of 284 review comments sampled from the original dataset using a different large language model, 
\emph{Ministral-8B-Instruct-2410}\footnote{\url{https://mistral.ai/}}. 
The same prompt from Section~\ref{sec:4.2} is adopted to ensure consistency. 
Evaluation metrics remain aligned with those defined in Sections~\ref{sec:4.2} and~\ref{sec:4.4}. 
For automated evaluation, we utilize \emph{gpt-4o-2024-11-20}\footnote{\url{https://openai.com/}} as the evaluation model to generate objective and standardized judgments across all dimensions.

\begin{figure*}
    \centering
    \includegraphics[width=0.98\linewidth]{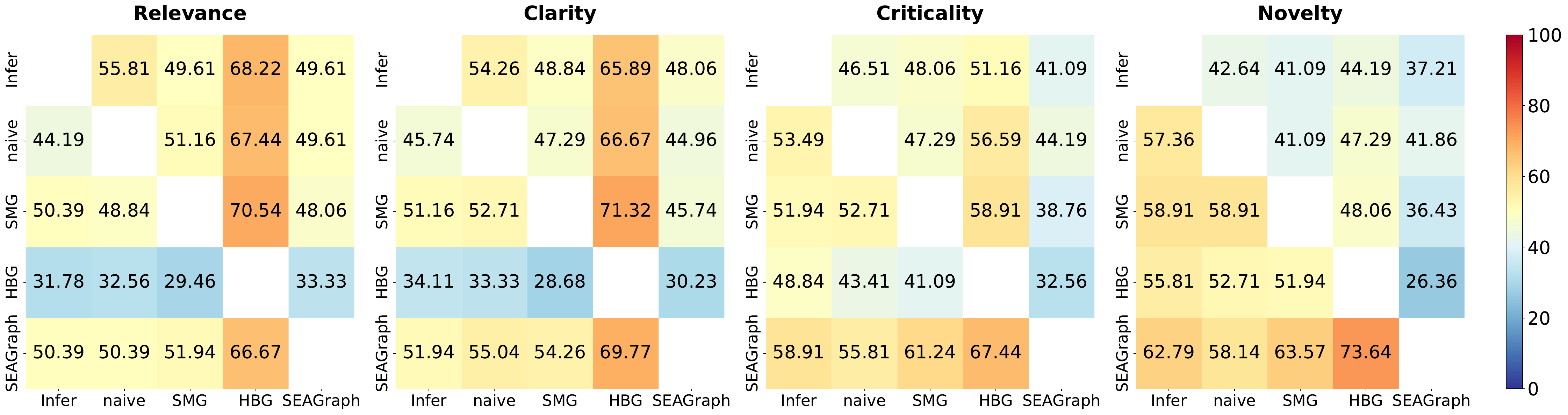}
    \caption{Automated evaluation results for Review Comments Understanding, measured by Win-Rate (\% $\uparrow$), based on comments with length $< 100$. }
    \label{fig:gpt_eval_100}
\end{figure*}

\begin{figure*}[ht]
    \centering
    \includegraphics[width=0.98\linewidth]{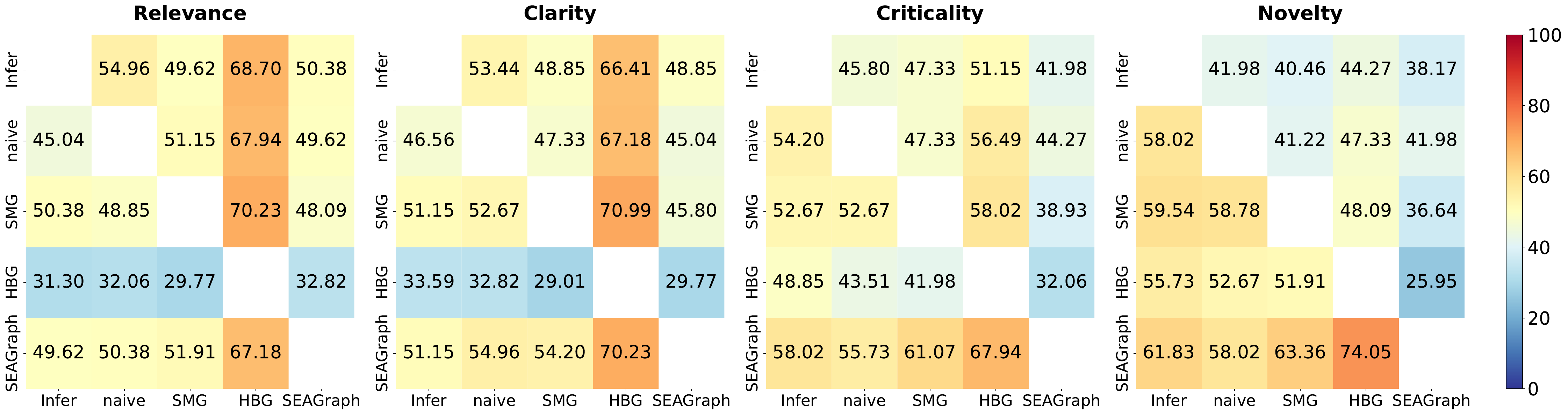}
    \caption{Automated evaluation results for Review Comments Understanding, measured by Win-Rate (\% $\uparrow$), based on comments with length in the range of $(100, 200)$.}
    \label{fig:gpt_eval_200}
\end{figure*}

\subsection{Automated Evaluation Results.}

In the automated evaluation, 
due to the excessive total text length generated by the five methods, we employ a pairwise ranking approach for assessment.
Figure~\ref{fig:gpt_eval_100} presents the results for review comments with a length of less than 100, while the results. 
The values in the heatmap represent the win rate of the method shown on the vertical axis over the method on the horizontal axis. 
From the figure, it can be observed that: 
(1) SEAGraph consistently performs as the optimal method across the four metrics.
(2) In terms of Criticality and Novelty metrics, 
\modelname~significantly outperforms other methods, indicating its ability to provide more innovative evidence and conduct deeper analysis for review comments. 
(3) In most cases, RAG-SMG consistently ranks second. 
Although it performs slightly worse than RAG-naive in terms of relevance, it surpasses RAG-naive across all other metrics.
This suggests that while RAG-SMG captures slightly less relevant aspects of evidence, its ability to make logical connections greatly enhances the reasoning power of LLM. 
These findings highlight the crucial role of modeling academic papers as semantic mind graphs
to capture the paper's underlying structure and logic.
(4) Notably, for novelty, RAG-HBG performs better than all methods except \modelname, as it introduces multiple perspectives beyond the reviewed paper.

Figure~\ref{fig:gpt_eval_200} shows the results for samples of all lengths.
From the figure, we can see that~\modelname~outperforms in most cases, only slightly falling short on the Relevance metric when compared to Infer. 
On the other hand, RAG-SMG lags slightly behind RAG-naive in terms of relevance, likely because longer review comments are already sufficiently detailed, leading to a minor disadvantage in argument relevance. 
However, both~\modelname~and RAG-SMG demonstrate superior performance on other metrics, proving that the evidence they provide are more effective and better support reasoning.

\subsection{Human Evaluation Results.}
In the human evaluation process, experts are invited to rank the review understanding results produced by five different methods based on six aforementioned metrics.
The results are summarized in Table~\ref{tab:human_eval_generate}, where the scores indicate the average ranking over all samples.
We can observe that:
(1) \modelname~achieves the highest performance across all metrics, underscoring the effectiveness of our framework.
(2) In terms of average rankings, 
RAG-SMG performs better than RAG-naive, which in turn outperforms DirectInfer.
These findings strongly support our motivation that logically retrieved content significantly improves the ability of LLMs to understand review comments.
(3) Although RAG-HBG performs poorly—mainly because its retrieved content consists only of background knowledge without incorporating the internal knowledge of the paper—\modelname\ enhances the LLM’s ability to understand the paper by integrating external background knowledge on top of RAG-SMG, enabling it to extract more meaningful information for reasoning.
Consequently,
constructing the semantic mind graph and the hierarchical background graph can provide valuable support for understanding paper reviews from different perspectives.

Overall, both automated and human evaluations yield consistent conclusions, 
providing strong evidence that \modelname~is capable of generating valuable insights for interpreting review comments.

\subsection{Analysis of Human Evaluation Consistency}
\label{app:human_constiency}
To assess the reliability and effectiveness of human evaluation in our study, we conduct a detailed analysis of NDCG@5 scores for Ministral-8B. 
NDCG@5 is computed based on independent human judgments for the model’s generated outputs. This metric quantifies how well annotators can distinguish and rank high-quality responses from lower-quality ones within the model’s output set. 
By examining NDCG@5 scores across multiple evaluation criteria, we aim to understand the overall reliability of human judgments and their sensitivity to meaningful variations in response quality.

The results, summarized in Table~\ref{ndcg_mini}, show the average NDCG@5 scores of Ministral-8B across six evaluation criteria: Relevance, Clarity, Critical Insight, Novelty, Persuasiveness, and Practicality. The consistently high scores across all dimensions suggest that human annotators were able to reliably differentiate response quality within the model’s generated outputs. This indicates strong agreement among annotators and provides evidence that the evaluation protocol captures stable and interpretable human judgments. Such consistency reinforces the credibility of our human evaluation setup and supports the reliability of conclusions drawn from it.

\input{table/ndcg_mini}

\section{Experts information for human evaluation}
\label{app:experts}
For human evaluation, we recruit 40 experts, including master’s and doctoral students from diverse academic backgrounds. 
Each review comment is independently evaluated by two experts to assess the quality of understanding.
All participants with prior experience in publishing peer-reviewed papers or serving as academic conference reviewers were compensated at a rate of \$10 per hour.

\input{table/generate_result}

\section{More Details of \modelname}

\begin{figure*}[ht]
    \centering 
\includegraphics[width=1.0\linewidth]{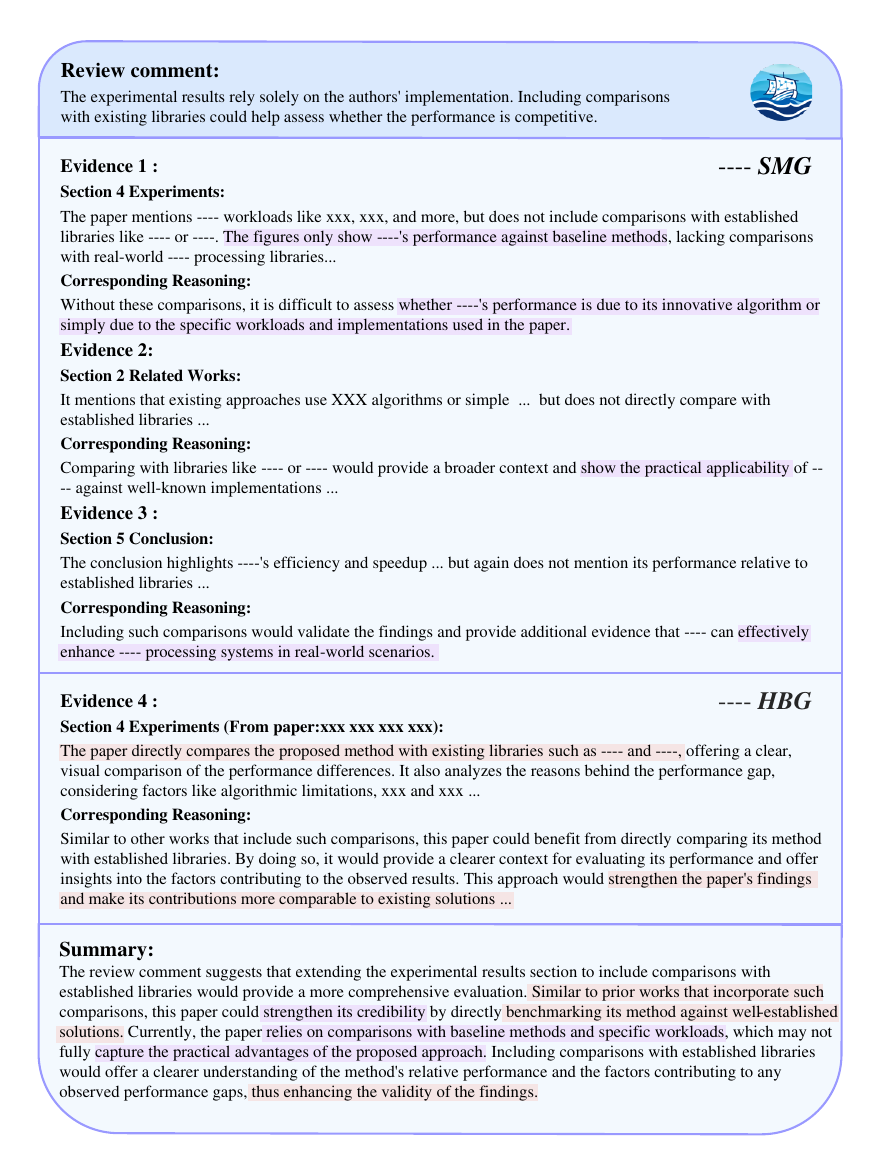}
    \caption{The example of~\modelname.}
    \label{fig:seagraph_exmaple}
\end{figure*}

\paragraph{Prompt.}
In Table~\ref{tab:generate_prompt}, 
we present the instruction designed to generate content for understanding review comments that conform to the specified format based on the retrieved content.
We require LLMs to output several evidence before the summary.

\paragraph{The Specific Meaning of Metrics.}
\label{specific_meaning_metrics}
The metrics for evaluating the generation of understanding review comments and the retrieval content are shown in Table~\ref{tab:metrics_generate} and Table~\ref{tab:metrics_retrievel}, respectively.

\section{The Generated Example of \modelname}
\label{app:generated_example_seagraph}
Figure~\ref{fig:seagraph_exmaple} shows an example of the explanation for a review comment generated by \modelname. For privacy concerns, both the review comment and the generated content have been processed. As shown in the figure, the review comment points out that the paper only conducts experiments in its own designed experimental settings, and suggests that comparing the proposed method with other libraries would help demonstrate its validity. \modelname~locates, through the retrieval of the semantic mind graph, that the paper mentions only part of the workloads in the ``Experiments'' and ``Related Work'' sections, highlighting that, although the method is effective, it does not compare with some of the libraries or algorithms mentioned. It also points out in the ``Conclusion'' that the method's validity could be verified by comparing it with more real-world applications. Additionally, \modelname, through the retrieval of hierarchical background knowledge, mentions other papers that have conducted such comparisons in their experiments. Finally, in the summary, \modelname~effectively consolidates the logic of the entire review comment, highlighting the missing experiments in the paper and referring to other papers' experimental settings. This example demonstrates \modelname’s ability to generate explanations for review comments by constructing two graphs and retrieving relevant chunks.

\section{The Example of SMG and HBG Construction}
\label{app:construct_example}
Taking SEAGraph (our paper) as an example, we showcase the structure of the constructed SMG and HBG of our paper. 
In a format similar to Figure~\ref{fig:smg}, we present the content in a linear fashion, displaying the section from Line 90 ``Building upon...'' to Line 105 ``that explains the reviewer's comments,'' along with the connected chunks, forming the full content of the article in Table~\ref{tab:example_smg}. 
Additionally, we highlight the different themes within the HBG and their corresponding papers in Table~\ref{tab:example_hbg}.
It becomes evident that the SMG and HBG we constructed are highly correlated with the review comments, which aligns with our experimental results.

\input{table/example_retrieve}

\input{table/example_smg}

\input{table/example_hbg}
\section{Retrieval Example of Semantic Mind Graph}
\label{app:retreiva_example}
To demonstrate the efficacy of the semantic mind graph, we conduct a comparative analysis between our proposed RAG-SMG approach and the baseline RAG-naive method. As illustrated in Table~\ref{tab:retrieve_example}, the RAG-naive system produces fragmented retrieval outcomes due to its exclusive reliance on maximizing semantic similarity without accounting for the paper's structural context. In contrast, RAG-SMG maintains contextual alignment with the review comment while simultaneously leveraging both the paper's structural coherence and semantic relationships. This dual consideration enables the generation of more cohesive retrieval results that systematically organize the review comment in a logical progression, transitioning from background information through contributions, evaluation protocols, and ultimately experimental conclusions.

\section{Hierarchical Background Graph Retrieval}
\label{app:hkeg_equation}
In the Hierarchical Background Graph, to make background knowledge more aligned with the reviewed paper and the review comments, 
we conduct a retrieval process for external knowledge based on the review comments and semantic mind subgraph identified in Section~\ref{sec:smg_retrieval}.
First, we compute the representation of the retrieved semantic mind subgraph $\bar{\mathcal{V}}$ using a pooling operation to obtain the subgraph representation.
First, we obtain the subgraph representation by applying a pooling operation to the node representations of the retrieved semantic mind subgraph.
\begin{equation}
    h_{\bar{\mathcal{V}}} = \frac{1}{|{\bar{\mathcal{V}}|}} \sum_{c_i \in \bar{\mathcal{V}}} h_{c_i} 
\end{equation}
On this basis, we design a three-layer retrieval framework, including theme-level, abstract-level, and chunk-level retrieval, to capture information at varying levels of granularity. 
The specific retrieval formulas are as follows:

\paragraph{(1) Theme level.}
\[
\text{score}(h_{t_i}) = \alpha_t \cdot \mathit{f}(h_{t_i},h_q) + \beta_t \cdot \mathit{f}(h_{t_i}, h_{S_s})
\]

\paragraph{(2) Abstract level.}
\begin{align*}
\text{score}(h_{a_i}) = \alpha_a \cdot \mathit{f}(h_{a_i}, h_q) 
&+ \beta_a \cdot \mathit{f}(h_{a_i}, h_{S_s}) \\
&+ \gamma_a \cdot \mathit{f}(h_{a_i}, h_{t_a})
\end{align*}

\paragraph{(3) Chunk level.}

\begin{align*}
\text{score}(h_{c_i}) 
& = \alpha_c \cdot \mathit{f}(h_{c_i}, h_q)  
+ \beta_c \cdot \mathit{f}(h_{c_i}, h_{S_s}) \\
&+ \gamma_c \cdot \mathit{f}(h_{c_i}, h_{t_a})
+ \delta_c \cdot \mathit{f}(h_{c_i},h_{a_c)}
\end{align*}

Here, $\alpha$, $\beta$, $\gamma$, and $\delta$ represent the hyperparameters, while $t_i$, $a_i$, and $c_i$ correspond to the text of the theme description, the abstract, and the chunk, respectively.
Additionally, $h_q$ denotes the embedding of the query.

\input{table/generate_prompt}
\input{table/metric_description_generate}
\input{table/metric_description_retrievel}

%% file: table/ndcg_mini.tex
\begin{table}[htbp]
  \centering
    \resizebox{0.65\linewidth}{!}{
  \begin{tabular}{lcc}
    \toprule
    Metric          & Ministral-8B \\
    \midrule
    Relevance       & 0.8761        \\
    Clarity         & 0.8600        \\
    Critical Insight  & 0.8785        \\
    Novelty           & 0.8616        \\
    Persuasiveness    & 0.8664        \\
    Practicality      & 0.8675        \\
    \bottomrule
  \end{tabular}
  }
  \caption{NDCG@5 of Ministral based on human evaluation.}
  \label{ndcg_mini}
\end{table}

%% file: table/generate_result.tex
\begin{table*}[t]
    \setlength\tabcolsep{3pt} 
    \renewcommand{\arraystretch}{1.2} 
    \centering
    \resizebox{0.85\linewidth}{!}{
    \begin{tabular}{l|cccccc|c}
    \toprule
    \textbf{Method}  & \textbf{Relevance} & \textbf{Clarity} & \textbf{Criticality} & \textbf{Novelty} & \textbf{Practicality} & \textbf{Persuasiveness} & \textbf{Avg.Rank} \\
    \midrule
    DirectInfer & 3.09 & 2.94 & 2.97 & 3.53 & 3.16 & \underline{2.63} & 3.05 \\
    RAG-naive  & 2.72 & \underline{2.78} & 3.22 & 2.91 & \underline{2.72} & 3.16 & 2.92 \\
    RAG-SMG & \underline{2.63} & 2.97 & \underline{2.50} & \underline{2.41} & 2.91 & 2.78 & \underline{2.70} \\
    RAG-HBG  & 4.41 & 4.13 & 4.25 & 4.22 & 4.16 & 4.31 & 4.25 \\
    SEAGraph  & \textbf{2.16} & \textbf{2.19} & \textbf{2.06} & \textbf{1.94} & \textbf{2.06} & \textbf{2.13} & \textbf{2.09} \\
    \bottomrule
    \end{tabular}
    }
    \caption{Human evaluation results for Review Comments Understanding (Rank-Based \textbf{$\downarrow$}). 
    We highlight the best score on each metric in \textbf{bold} and the runner-up score with an \underline{underline}.
    }
     \label{tab:human_eval_generate}
\end{table*}

%% file: table/example_retrieve.tex
\begin{table*}[htbp]
\centering
\begin{tabular}{|p{0.45\textwidth}|p{0.45\textwidth}|}
\hline
\multicolumn{2}{
  >{\columncolor{gray!20}}p{\dimexpr0.9\textwidth + 2\tabcolsep}%
}{\textbf{Review Comment: The scenarios and tasks focused in this paper are meaningful and may facilitate the peer-review process.}} \\
\hline
\textbf{RAG-naive} & \textbf{RAG-SMG} \\
\hline

\# Expanding on this research, Jin et al. (2024) employ LLMs to simulate the entire review process, thereby revealing the impact of various factors on academic evaluation... However, both human reviews and generated reviews may suffer from issues such as ambiguity or brevity, causing confusion for authors. Our work aims to leverage LLMs to understand the intent of review comments, thereby assisting authors in polishing their papers. \textbf{(excerpt from 2 Related Work Large Language Models in Peer Review)}
\makecell[l]{\\}

\# In recent years, the number of academic publications has grown exponentially, creating a vast "sea of papers" Traditionally, authors rely on the peer review process to receive feedback on their manuscripts. However, the review cycle typically requires several months or even longer, which is time-consuming [1]... the statement that "the method is limited" is very vague without any details provided. \textbf{(excerpt from 1 Introduction)}
\makecell[l]{\\}

\# SEAGraph is designed to assist authors in comprehending review comments during the peer review process, with a particular emphasis on the review-comment understanding stage. \textbf{(excerpt from 5 Conclusion)}
\makecell[l]{\\}

\# Baseline Methods. To validate the effectiveness of SEAGraph in terms of graph construction and retrieval, we compare it with the following two categories of baseline methods: (1) Direct inference methods: ... **RAG-SMG** utilizes only the construction and retrieval of the **S**emantic **M**ind **G**raph; **RAG-HBG** relies solely the construction and retrieval of the **H**ierarchical **B**ackground **G**raph. \textbf{(excerpt from 4 Experiments Experimental Setup)}
\makecell[l]{\\}

...

&
\# In recent years, the number of academic publications has grown exponentially, creating a vast "sea of papers" Traditionally, authors rely on the peer review process to receive feedback on their manuscripts However, the review cycle typically requires several months or even longer, which is time-consuming [1]. Meanwhile, the large volume of submissions results in uncertain review qualities [1], often resulting in ambiguous or overly brief comments that are challenging to explain [14]. For example, the statement that "the method is limited" is very vague without any details provided. \textbf{(excerpt from 1 Introduction)}
\makecell[l]{\\}

\#\# In this paper, we present SEAGraph, a novel framework designed to bridge the gap between reviewers' comments and authors' understanding...
SEAGraph not only enhances the clarity of reviews but also empowers authors to understand reviewer concerns more effectively, improving the quality of academic publications. \textbf{(excerpt from 5 Conclusion)}
\makecell[l]{\\}

\#\#\# Evaluation Protocol. Since the task of review comments understanding lacks a definitive ground truth and exhibits significant diversity in generated content, we design two evaluation methods: \textit{human evaluation} and \textit{automated evaluation}. For human evaluation, ..., we employ gpt-4o-2024-11-203 as the evaluation model to provide objective judgments for the task. \textbf{(excerpt from 4 Experiments Experimental Setup)}
\makecell[l]{\\}

\#\#\# From the figure, it can be observed that:
(1) SEAGraph consistently outperforms all baseline methods across all four evaluation dimensions,... Overall, SEAGraph is capable of
generating valuable insights for interpreting review comments.  \textbf{(excerpt from 4 Experiments Main experiments)}
\makecell[l]{\\}

...

\\
\hline
\end{tabular}
\label{tab:retrieve_example}
\caption{Comparison of retrieval results between RAG-naive and RAG-SMG (with ‘…’ indicating omitted intermediate content).}
\end{table*}

%% file: table/example_smg.tex
\begin{table*}[ht]
\centering
\footnotesize
\phantomsection
\begin{tcolorbox}[colback=white!95!gray,colframe=gray!50!black,rounded corners,label={scale-depression}, title={Constructed Example of SMG.
}]
\begin{lstlisting}[breaklines=true, xleftmargin=0pt, breakindent=0pt, columns=fullflexible, mathescape]
- SEAGraph: Unveiling the Whole Story of Paper Review Comments (excerpt from Title)
- Peer review, as ... (excerpt from Abstract)
- Another alternative approach is using RAG ... which splits lengthy texts into discrete chunks and hierarchically connects them, has inspired new directions. (excerpt from 1 Introduction)
- Similarly, ... hierarchical background graph. (excerpt from 1 Introduction)
- Building upon the principles of the mind map ... we introduce the semantic mind graph to facilitate deeper semantic connections and organization of key knowledge points. In addition, the hierarchical background graph connects various related papers based on the themes of the paper, thereby simulating its research context. After the construction of the two graphs, we design a tailored retrieval method to extract the most relevant content from both graphs in response to each review comment. The extracted content is subsequently fed into LLMs to generate coherent and logical arguments that explain the reviewer's comments. (excerpt from 1 Introduction)
- We construct a semantic mind graph and a hierarchical background graph for a paper, capturing its deep semantics and related domain knowledge. (excerpt from 1 Introduction)
- Paper Processing. ... respectively. Review Comments Extraction. (excerpt from 3 SEAGraph Review Comments Understanding Task)
- Our goal is to retrieve subgraphs from ... By generating a logical chain, SEAGraph enables authors to better understand reviewers' perspectives and proceed with subsequent research more effectively. (excerpt from 3 SEAGraph Review Comments Understanding Task)
- A paper is structurally organized into different sections, while key points of the paper may be scattered across various parts. The entire paper can be structured like a mind graph, where content progressively branches out from different paragraphs. Our goal is to construct a semantic mind graph to model the writing logic of a paper. (excerpt from 3 SEAGraph Semantic Mind Graph Construction)
- We next detail the main steps. Paper Chunking ... whether adjacent sentences should be merged into chunks. (excerpt from 3 SEAGraph Semantic Mind Graph Construction)
- Figure 2: ... the embedding similarity between (s_{i}) and (c_{\text{current}}). (excerpt from 3 SEAGraph Semantic Mind Graph Construction)
- Further, the authors may not ... through practical evaluations. (excerpt from 3 SEAGraph Semantic Mind Graph Construction)
- Finally, the paper is transformed into a semantic mind graph ... semantic similarity. (excerpt from 3 SEAGraph Semantic Mind Graph Construction)
- To effectively review ... from the "Reference" section. (excerpt from 3 SEAGraph Hierarchical Background Graph Construction)
- Now, for a reviewed paper, we ... offering a deeper insight into the paper's structure and details. (excerpt from 3 SEAGraph Hierarchical Background Graph Construction)
- We next introduce retrieving the semantic mind graph based on review comments and obtaining the relevant supporting texts. Given a review comment as a query, we first calculate the probability distribution of the query over the semantic mind graph by calculating the textual similarity between the query and each chunk node (c_{j}): (excerpt from 3 SEAGraph Semantic Mind Graph Retrieval)
- To further mine the background knowledge related to a paper, we conduct ... aiming to extract themes related to a review comment. (excerpt from 3 SEAGraph Hierarchical Background Graph Retrieval)
- For example, if a reviewer questions ... (excerpt from 3 SEAGraph Hierarchical Background Graph Retrieval)
- Finally, we step into the chunk-level retrieval. ... we feed the corresponding text along with the query into LLM to generate an explanation for the review comment. (excerpt from 3 SEAGraph Hierarchical Background Graph Retrieval)
- Baseline Methods. To validate ... and retrieval of the Hierarchical Background Graph. (excerpt from 4 Experiments Experimental Setup)
- Overall, both human and automated evaluations yield consistent conclusions, providing strong evidence that SEAGraph is capable of generating valuable insights for interpreting review comments. (excerpt from 4 Experiments Main experiments)
- In this section, ... The results are shown in Figure 5. (excerpt from 4 Experiments Key Concerns in Reviews)
- In this paper, we present SEAGraph, ... improving the quality of academic publications. (excerpt from 5 Conclusion)
- In the Hierarchical Background Graph, to make ... semantic mind subgraph. (excerpt from Appendix D Hierarchical Background Graph Retrieval)

\end{lstlisting}
\end{tcolorbox}
\caption{Constructed Example of SMG.
The '...' in the following text represents content from the original document, which has been omitted due to length constraints.
}
\label{tab:example_smg}
\end{table*}

%% file: table/example_hbg.tex
\begin{table*}[ht]
\centering
\small
\phantomsection
\begin{tcolorbox}[colback=white!95!gray,colframe=gray!50!black,rounded corners,label={scale-depression}, title={Constructed Example of HBG.
}]
\begin{lstlisting}[breaklines=true, xleftmargin=0pt, breakindent=0pt, columns=fullflexible, mathescape]
Theme 1 : Knowledge Retrieval
- Title: Query Rewriting for Retrieval-Augmented Large Language Models
- Title: Enhancing Structured-Data Retrieval with GraphRAG: Soccer Data Case Study
- Title: Retrieval-Augmented Generation for Knowledge-Intensive NLP Tasks
- Title: Retrieval-Augmented Generation for Large Language Models: A Survey
- Title: From Local to Global: A Graph RAG Approach to Query-Focused Summarization
- Title*: Survey on Factuality in Large Language Models: Knowledge, Retrieval and Domain-Specificity
- Title*: Context Recovery and Knowledge Retrieval: A Novel Two-Stream Framework for Video Anomaly Detection

Theme 2 : Graph-Based Retrieval
- Title: Enhancing Structured-Data Retrieval with GraphRAG: Soccer Data Case Study
- Title: Generate rather than Retrieve: Large Language Models are Strong Context Generators
- Title: LEGO-GraphRAG: Modularizing Graph-based Retrieval-Augmented Generation for Design Space Exploration
Title*: Graph Retrieval-Augmented Generation: A Survey

Theme 3 : Natural Language Processing (NLP)
- Title: Query Rewriting for Retrieval-Augmented Large Language Models
- Title: Enhancing Structured-Data Retrieval with GraphRAG: Soccer Data Case Study
- Title: Retrieval-Augmented Generation for Knowledge-Intensive NLP Tasks
- Title: AutoGen: Enabling Next-Gen LLM Applications via Multi-Agent Conversation
- Title: A Survey of Large Language Models
- Title*: Demystifying the Role of Natural Language Processing (NLP) in Smart City Applications: Background, Motivation, Recent Advances, and Future Research Directions

Theme 4 : Peer Review and Academic Reliability
- Title: Are We There Yet? Revealing the Risks of Utilizing Large Language Models in Scholarly Peer Review
- Title: AcademicGPT: Empowering Academic Research
- Title: Can large language models provide useful feedback on research papers? A large-scale empirical analysis.
- Title: Is Your Paper Being Reviewed by an LLM? Investigating AI Text Detectability in Peer Review
- Title: Automated Peer Reviewing in Paper SEA: Standardization, Evaluation, and Analysis
- Title: Are peer-reviews of grant proposals reliable? An analysis of Economic and Social Research Council (ESRC) funding applications

Theme 5 : Multi-Agent Conversations and Interaction
- Title: AutoGen: Enabling Next-Gen LLM Applications via Multi-Agent Conversation
- Title: AgentReview: Exploring Peer Review Dynamics with LLM Agents
- Title: ReviewerGPT? An Exploratory Study on Using Large Language Models for Paper Reviewing
- Title: From Local to Global: A Graph RAG Approach to Query-Focused Summarization
- Title*: ChoiceMates: Supporting Unfamiliar Online Decision-Making with Multi-Agent Conversational Interactions
- Title*: MuMA-ToM: Multi-modal Multi-Agent Theory of Mind

\end{lstlisting}
\end{tcolorbox}
\caption{Constructed Example of HBG. 
An asterisk indicates the latest or most popular articles found on Google Scholar based on the theme, while other papers are those included in the Related Work section.}
\label{tab:example_hbg}
\end{table*}

%% file: table/generate_prompt.tex
\begin{table*}[ht]
\centering
\phantomsection
\begin{tcolorbox}[colback=white!95!gray,colframe=gray!50!black,rounded corners,label={scale-depression}, title={Prompt of the task of Review Comment Understanding. }]
\begin{lstlisting}[breaklines=true, xleftmargin=0pt, breakindent=0pt, columns=fullflexible, mathescape]
You are an experienced researcher with strong logical thinking and excellent reasoning skills. You will receive a paper along with a corresponding review comment. We have provided the key sections of the paper and the critical content from related work. The review comment is found between <REVIEW> and </REVIEW>, the key sections of the paper are between <PAPER HIGHLIGHTS> and </PAPER HIGHLIGHTS>, and the related work is found between <RELATED WORK> and </RELATED WORK>. 

Your task is to systematically find supporting evidence and construct a complete logical chain, thereby building a full reasoning chain to clarify why the reviewer made this comment. 

Please structure the reasoning chain as follows:

- **Evidence 1 (specific section) **: 
    <corresponding reasoning>
    
- **Evidence 2 (specific section)**: 
    <corresponding reasoning>

- ... (more evidence if available)

- **Summary**:
<Logical reasoning based on evidence explaining the basis for the review comment.>

\end{lstlisting}
\end{tcolorbox}
\caption{Instruction for generating understanding content for review comments.}
\label{tab:generate_prompt}
\end{table*}

%% file: table/metric_description_generate.tex
\begin{table*}[ht]
\centering
\renewcommand{\arraystretch}{1.3}
\setlength{\tabcolsep}{10pt}
\begin{tabular}{|p{4cm}|p{10cm}|}
\hline
\rowcolor[HTML]{EFEFEF} 
\textbf{Metric} & \textbf{Description} \\ \hline
\textbf{Relevance} & Is the evidence in the supplementary information highly relevant and closely aligned with the reviewers’ comments? \\ \hline
\textbf{Clarity} &  Is the supplementary information clearly articulated and easy to understand? Does it effectively explain the reviewers’ viewpoints and the supporting arguments? \\ \hline
\textbf{Criticality} & Does the supplementary information provide an in-depth analysis and reflection on the reviewers’ feedback? Does it identify any limitations in the feedback and offer reasonable suggestions for improvement? \\ \hline
\textbf{Novelty} & Does the supplementary information present unique insights or new evidence not mentioned in the original review, thereby enriching the depth and breadth of the content? \\ \hline
\textbf{Persuasiveness} & Does the summary present the evidence in a compelling manner, demonstrating logical reasoning and effectively persuading the reviews of the core ideas with clarity and coherence? \\ \hline
\textbf{Practicality} & Does the supplementary information provide direct assistance to the author? \\ \hline
\end{tabular}
\caption{Metrics for Evaluating Supplementary Information.}
\label{tab:metrics_generate}
\end{table*}

%% file: table/metric_description_retrievel.tex
\begin{table*}[ht]
\centering
\renewcommand{\arraystretch}{1.3}
\setlength{\tabcolsep}{10pt}
\begin{tabular}{|p{3.5cm}|p{11cm}|}
\hline
\rowcolor[HTML]{EFEFEF} 
\textbf{Metric} & \textbf{Description} \\ \hline
\textbf{Relevance} & How closely does the retrieved information relate to the topic of the paper review or the content of the paper? \\ \hline
\textbf{Specificity} & Is the retrieved information detailed and specific? Can it effectively supplement the review content or provide new insights? \\ \hline
\textbf{Novelty} & Does the retrieved information offer a new perspective or provide supportive evidence not mentioned in the review? \\ \hline
\textbf{Logic} & Is the retrieved information consistent with the review content and the overall logic of the paper? \\ \hline
\textbf{Explainability} & Can it effectively address the issues mentioned in the review or provide theoretical foundations or case studies to back up the review’s arguments? \\ \hline
\end{tabular}
\caption{Metrics for Evaluating Retrieved Information.}
\label{tab:metrics_retrievel}
\end{table*}